\documentclass[10pt,twocolumn,letterpaper]{article}
 
 \usepackage{cvpr}
 \usepackage{times}
 \usepackage{epsfig}
 \usepackage{graphicx}
 \usepackage{amsmath}
 \usepackage{amssymb}
 \usepackage{color}
 \usepackage{svg}
 \usepackage{floatrow}
 \usepackage{blindtext}
 \usepackage{pstricks-add}
 \usepackage{subfig}
 \usepackage{multirow}
 \usepackage{xcolor,colortbl}
 \usepackage{array}
 \usepackage{mathtools}
\usepackage{enumitem}
\usepackage[pagebackref=true,breaklinks=true,letterpaper=true,colorlinks,bookmarks=false]{hyperref}
\usepackage{multirow}
\usepackage{hhline}
\usepackage[font=footnotesize,labelfont=bf]{caption}

 \newcolumntype{P}[1]{>{\centering\arraybackslash}p{#1}}
 \newcolumntype{M}[1]{>{\centering\arraybackslash}m{#1}}
 
 \definecolor{FEI}{rgb}{1,0,0}
 \definecolor{AMIR}{rgb}{0,0.5,0}
 \definecolor{JERRY}{rgb}{0,0,1}

 \usepackage[breaklinks=true,bookmarks=false]{hyperref}
 
 \cvprfinalcopy 
 
 \def\cvprPaperID{744} 

 \ifcvprfinal\fi
\makeatletter
\let\suppsavedmaketitle\maketitle
\let\suppsavedatmaketitle\@maketitle
\let\suppsavedtitle\title
\let\suppsavedauthor\author
\let\suppsavedand\and
\let\suppsavedthanks\thanks
\let\suppsaveddate\date
\makeatother
\DeclareRobustCommand{\suppref}[1]{{\hypersetup{linkcolor=magenta}\hyperlink{supplement.start}{#1}}}
\makeatletter\def\@fnsymbol#1{\ensuremath{\ifcase#1\or *\or \dagger\or \ddagger\or \mathsection\or \mathparagraph\or \|\or **\or \dagger\dagger\or \ddagger\ddagger \else\@ctrerr\fi}}\makeatother
\begin{document}

\title{Gibson Env: Real-World Perception for Embodied Agents\vspace{-1mm}}

\author{Fei Xia\thanks{Authors contributed equally.}$_{1}$ \hspace{1mm} \;\;Amir R. Zamir\footnotemark[1]$_{1,2}$ \hspace{1mm} \;\;Zhi-Yang He\footnotemark[1]$_{1}$  \;\; Alexander Sax$_{1}$  \;\; Jitendra Malik$_{2}$ \;\; Silvio Savarese$_{1}$\vspace{5pt}\\ 
	$^1$ Stanford University  \;\;  
	$^2$ University of California, Berkeley\vspace{5pt}\\ 
	{\url{http://gibsonenv.stanford.edu/}\vspace{0pt}}
}

\thispagestyle{empty}

\twocolumn [{%
	\renewcommand\twocolumn[1][]{#1}%
	\maketitle
		\vspace{-8mm}
		\includegraphics[width=.99\linewidth]{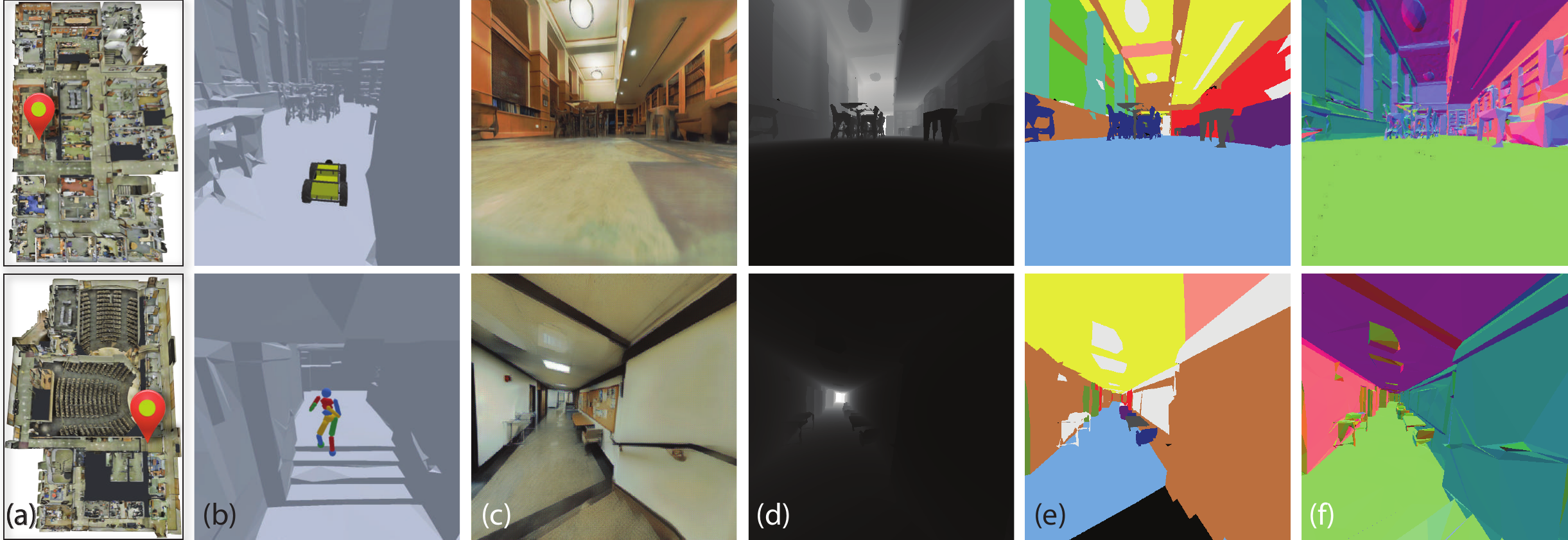}
		\captionof{figure}{\footnotesize{\textbf{Two agents in \emph{Gibson Environment} for real-world perception}. The agent is active, embodied, and subject to constraints of physics and space (a,b). It receives a constant stream of visual observations as if it had an on-board camera (c). It can also receive additional modalities, e.g. depth, semantic labels, or normals (d,e,f). The visual observations are from real-world rather than an artificially designed space.\vspace{-2mm}}}%
		\label{fig:pull}
	\vspace{4mm}
}]

\begin{abstract}
\vspace{-1mm}
Developing visual perception models for active agents and sensorimotor control are cumbersome to be done in the physical world, as existing algorithms are too slow to efficiently learn in real-time and robots are fragile and costly. This has given rise to learning-in-simulation which consequently casts a question on whether the results transfer to real-world. In this paper, we are concerned with the problem of developing \textbf{real-world perception} for \textbf{active agents}, propose \textbf{Gibson Virtual Environment}
\footnote{Named after \textbf{JJ Gibson}, the author of \emph{Ecological Approach to Visual Perception}, 1979. ``We must perceive in order to move, but we must also move in order to perceive" -- JJ Gibson~\cite{gibson2013ecological} \\ ${  \indent }^*$Authors contributed equally.} 
for this purpose, and showcase sample perceptual tasks learned therein.
Gibson is based on virtualizing real spaces, rather than using artificially designed ones, and currently includes over 1400 floor spaces from 572 full buildings.
The main characteristics of Gibson are:
I. being from the real-world and reflecting its semantic complexity, 
II. having an internal synthesis mechanism, ``Goggles", enabling deploying the trained models in real-world without needing domain adaptation, 
III. embodiment of agents and making them subject to constraints of physics and space.


\end{abstract}

\section{Introduction}
\label{sec:intro}
\vspace{-1mm}
We would like our robotic agents to have compound perceptual and physical capabilities: a drone that autonomously surveys buildings, a robot that rapidly finds victims in a disaster area, or one that safely delivers our packages, just to name a few. 
Apart from the application perspective, the findings supportive of a close relationship between visual perception and being physically active are prevalent on various fronts: evolutionary and computational biologists have hypothesized a key role for intermixing perception and locomotion in development of complex behaviors and species~\cite{parker2011origin, wolpert2000computational, churchland1994critique}; neuroscientists have extensively argued for a hand in hand relationship between developing perception and being active~\cite{smith2005development, held1963movement}; pioneer roboticists have similarly advocated entanglement of the two~\cite{brooks1990elephants, brooks1991intelligence}. 
This all calls for developing principled perception models specifically with active agents in mind. 

By perceptual active agent, we are generally referring to an agent that receives a visual observation from the environment and accordingly effectuates a set of actions which can lead a physical change in the environment ($\sim$manipulation) and/or the agent's own particulars ($\sim$locomotion).  
Developing such perceptual agents entails the questions of \emph{how} and \emph{where} to do so. 

\vspace{2mm}
On the \emph{how} front, the problem has been the focus of a broad set of topics for decades, from classical control~\cite{pomerleau1989alvinn, boyd1994linear, khatib1987unified} to more recently sensorimotor control~\cite{finn2016deep, levine2016end, levine2016learning, abbeel2010autonomous}, reinforcement learning~\cite{abbeel2007application, schulman2015trust, schulman2017proximal}, acting by prediction~\cite{dosovitskiy2016learning}, imitation learning~\cite{codevilla2017end}, and other concepts~\cite{mottaghi2016newtonian, zhu2017target,wu2015galileo,wu2016physics}. These methods generally assume a sensory observation from environment is given and subsequently devise one or a series of actions to perform a task.

A key question is \emph{where} this sensory observation should come from. 
Conventional computer vision datasets~\cite{everingham2010pascal,deng2009imagenet, lin2014microsoft} are passive and static, and consequently, lacking for this purpose. Learning in the physical world, though not impossible~\cite{gupta2016supersizing, agrawal2016learning, levine2016learning, pinto2016supersizing}, is not the ideal scenario. It would bound the learning speed to real-time, incur substantial logistical cost if massively parallelized, and discount rare yet important occurrences. Robots are also often costly and fragile. 
This has led to popularity of learning-in-simulation with a fruitful history going back to decades ago~\cite{pomerleau1989alvinn,langley1981simulated,carbonell1986world} and remaining an active topic today. The primary questions around this option are naturally around \emph{generalization from simulation to real-world}: how to ensure I. the \emph{semantic complexity} of the simulated environment is a good enough replica of the intricate real-world, and II. the \emph{rendered visual observation} in simulation is close enough to what a camera in real-world would capture (photorealism). 

We attempt to address some of these concerns and propose Gibson, a virtual environment for training and testing \emph{real-world} perceptual agents. An arbitrary agent, e.g. a humanoid or a car (see Fig.~\ref{fig:pull}) can be imported, it will be then embodied (i.e. contained by its physical body) and placed in a large and diverse set of real spaces. The agent is subject to constraints of space and physics (e.g. collision, gravity) through integration with a physics engine, but can freely perform any mobility task as long as the constraints are satisfied. Gibson provides a stream of visual observation from arbitrary viewpoints as if the agent had an on-board camera. Our novel rendering engine operates notably faster than real-time and works given sparsely scanned spaces, e.g. 1 panorama per 5-10 $m^2$. 

The main goal of Gibson is to facilitate transferring the models trained therein to real-world, i.e. holding up the results when the stream of images switches to come from a real camera rather than Gibson's rendering engine. This is done by: 
first, resorting to the world itself to represent its own semantic complexity~\cite{simon1996sciences, brooks1990elephants} and forming the environment based off of scanned real spaces, rather than artificial ones~\cite{song2016ssc,kempka2016vizdoom,johnson2016malmo}. 
Second, embedding a mechanism to dissolve differences between Gibson's renderings and what a real camera would produce. As a result, an image coming from a real camera vs the corresponding one from Gibson's rendering engine look statistically indistinguishable to the agent, and hence, closing the (perceptual) gap. This is done by employing a neural network based rendering approach which jointly trains a network for making renderings look more like real images (forward function) as well as a network which makes real images look like renderings (backward function). The two functions are trained to produce equal outputs, thus bridging the two domains. The backward function resembles deployment-time \emph{corrective glasses} for the agent, so we call it \emph{Goggles}. 

Finally, we showcase a set of active perceptual tasks (local planning for obstacle avoidance, distant navigation, visual stair climbing) learned in Gibson. Our focus in this paper is on the vision aspect only. The statements should not be viewed to be necessarily generalizable to other aspects of learning in virtual environments, e.g. physics simulation.


Gibson Environment and our software stack are available to public for research purposes at \href{http://gibsonenv.stanford.edu/}{http://gibsonenv.stanford.edu/}. Visualizations of Gibson space database can be seen \href{http://gibsonenv.stanford.edu/database/}{here}.




\section{Related Work}
\label{sec:related}

\textbf{Active Agents and Control:}
As discussed in Sec.\ref{sec:intro}, operating and controlling active agents have been the focus of a massive body of work. 
A large portion of them are non-learning based~\cite{khatib1987unified, desai2001modeling,khatib1986real}, while recent methods have attempted learning visuomotor policies end-to-end~\cite{zhu2017target, levine2016end} taking advantage of imitation learning~\cite{ross2011reduction}, reinforcement learning~\cite{schulman2017proximal, heess2017emergence,schulman2015trust, heess2017emergence, abbeel2010autonomous, abbeel2007application}, acting by prediction~\cite{dosovitskiy2016learning} or self-supervision~\cite{gupta2016supersizing,pinto2016supersizing,dosovitskiy2016learning,pathak2017curiosity, hirose2018gonet}. These methods are all potential users of (ours and other) virtual environments.

\textbf{Virtual Environments for Learning:}
Conventionally vision is learned in static datasets~\cite{everingham2010pascal,deng2009imagenet, lin2014microsoft} which are of limited use when it comes to active agent. Similarly, video datasets~\cite{laptev2008learning,rodriguez2008action,xu2016end} are pre-recorded and thus passive. 
Virtual environments have been a remedy for this, classically~\cite{pomerleau1989alvinn} and today~\cite{zhu2017target, Gaidon:Virtual:CVPR2016, dosovitskiy2017carla, airsim2017fsr, jiang2017configurable, gupta2017cognitive, bellemare2013arcade, mattersim, ros2016synthia,ammirato2017dataset,wu2018building}. Computer games, e.g. Minecraft~\cite{johnson2016malmo}, Doom~\cite{kempka2016vizdoom} and GTA5~\cite{richter2016playing} have been adapted for training and benchmarking learning algorithms. While these simulators are deemed reasonably effective for certain planning or control tasks, the majority of them are of limited use for perception and suffer from oversimplification of the visual world due to using synthetic underlying databases and/or rendering pipeline deficiencies. Gibson addresses some of such concerns by striving to target perception in real-world via using real spaces as its base, a custom neural view synthesizer, and a baked-in adaption mechanism, \emph{Goggles}.

\begin{figure*}
    \vspace{-0.6cm}		
	\includegraphics[width=1\linewidth]{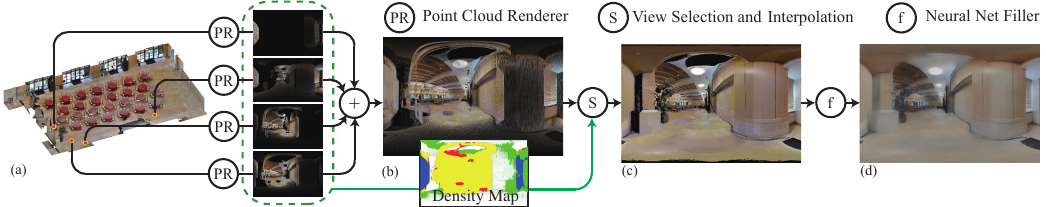}
    \vspace{-0.6cm}		
	\caption{{\textbf{Overview of our view synthesis pipeline.}} The input is a sparse set of RGB-D Panoramas with their global camera pose.  (a,b) Each RGB-D panorama is projected to the target camera pose and rendered.  (b) View Selection determines from which panorama each target pixel should be picked, favoring panoramas that provide denser pixels for each region. (c) The pixels are selected and local gaps are interpolated with bilinear sampling. (d) A neural network ($f$) takes in the interpolated image and fills in the dis-occluded regions and fixes artifacts.\vspace{-5mm}}
	\label{fig:vs}
\end{figure*}

\textbf{Domain Adaptation and Transferring to Real-World:}  
With popularity of simulators, different approaches for domain adaption for transferring the results to real world has been investigated~\cite{blitzer2006domain,daume2009frustratingly,sun2016return, saenko2010adapting,tobin2017domain,wulfmeier2017addressing}, e.g. via domain randomization~\cite{sadeghi2016rl, tobin2017domain} or forming joint spaces~\cite{sener2016learning}. Our approach is relatively simple and makes use of the fact that, in our case, large amounts of paired data for target-source domains are available enabling us to train forward and backward models to form a joint space. This makes us a baked-in mechanism in our environment for adaption, minimizing the need for additional and custom adaptation.

\textbf{View Synthesis and Image-Based Rendering:} 
Rendering novel views of objects and scenes is one of the classic problems in vision and graphics~\cite{seitz1996view, shechtman2010regenerative, suwajanakorn2014total,chen1993view,levoy1996light}. A number of relevantly recent methods have employed neural networks in a rendering pipeline, e.g. via an encoder-decoder like architecture that directly renders pixels~\cite{dosovitskiy2015learning,kulkarni2015deep, tatarchenko2016multi} or predicts a flow map for pixels~\cite{zhou2016view}. When some from of 3D information, e.g. depth, is available in the input~\cite{hartley2003multiple,mark1997post,chang1999ldi, shade1998layered}, the pipeline can make use of geometric approaches to be more robust to large viewpoint changes and implausible deformations. Further, when multiple images in the input are available, a smart selection mechanism (often referred to as Image Based Rendering) can help with lighting inconsistencies and handling more difficult and non lambertian surfaces~\cite{hedman2016scalable,ortiz2015bayesian, waechter2014let}, compared to rendering from a textured mesh or as such entirely geometric methods. Our approach is a combination of above in which we geometrically render a base image for the target view, but resort to a neural network to correct artifacts and fill in the dis-occluded areas, along with jointly training a backward function for mapping real images onto the synthesized one.


\section{Real-World Perceptual Environment}
\label{sec:system}


Gibson includes a neural network based view synthesis (described in Sec.~\ref{subsec:viewsyn}) and a physics engine (described in Sec.~\ref{subsec:physics}). The underlying scene database and integrated agents are explained in sections~\ref{subsec:data} and~\ref{subsec:physics}, respectively. 


\subsection{Gibson Database of Spaces}
\label{subsec:data}

Gibson's underlying database of spaces includes 572 full buildings composed of 1447 floors covering a total area of 211k $m^2$. Each space has a set of RGB panoramas with global camera poses and reconstructed 3D meshes. The base format of the data is similar to 2D-3D-Semantics dataset~\cite{2017arXiv170201105A}, but is more diverse and includes 2 orders of magnitude more spaces. Various 2D, 3D, and video visualizations of each space in Gibson database can be accessed \href{http://gibsonenv.stanford.edu/database/}{here}.
This dataset is released in asset files of Gibson\footnote{Stanford AI lab has the copyright to all models.}.


We have also integrated 2D-3D-Semantics dataset~\cite{2017arXiv170201105A} and Matterport3D~\cite{chang2017matterport3d} in Gibson for optional use.



\subsection{View Synthesis}
\label{subsec:viewsyn}
Our view synthesis module takes a sparse set of RGB-D panoramas in the input and renders a panorama from an arbitrary novel viewpoint. A `view' is a 6D camera pose of $x,y,z$ Cartesian coordinates and roll, pitch, yaw angles, denoted as $\theta, \phi, \gamma$. An overview of our view synthesis pipeline can be seen in Fig.~\ref{fig:vs}. It is composed of a geometric point cloud rendering followed by a neural network to fix artifacts and fill in the dis-occluded areas, jointly trained with a backward function. Each step is described below:

{\bf Geometric Point Cloud Rendering}. 
Scans of real spaces include sparsely captured images, leading to a sparse set of sampled lightings from the scene. The quality of sensory depth and 3D meshes are also limited by 3D reconstruction algorithms or scanning devices. Reflective surfaces or small objects are often poorly reconstructed or entirely missing. All these prevent simply rendering from textured meshes to be a sufficient approach to view synthesis. 

We instead adopt a two-stage approach, with the first stage being geometrically rendering point clouds: the given RGB-D panoramas are transformed into point clouds and each pixel is projected from equirectangular coordinates to Cartesian coordinates. For the desired target view $v_j = (x_j, y_j, z_j, \theta_j, \phi_j, \gamma_j)$, we choose the nearest $k$ views in the scene database, denoted as $v_{j,1}, v_{j,2},\dots, v_{j,k}$. For each view $v_{j,i}$, we transform the point cloud from $v_{j,i}$ coordinate to $v_{j}$ coordinate with a rigid body transformation and project the point cloud onto an equirectangular image. The pixels may open up and show a gap in-between, when rendered from the target view. Hence, the pixels that are supposed to be occluded may become visible through the gaps. To filter them out, we render an equirectangular depth as seen from the target view $v_j$ since we have the full reconstruction of the space. We then do a depth test and filter out the pixels with a difference $>0.1m$ in their depth from the corresponding point in the target equirectangular depth. We now have sparse RGB points projected in equirectangulars for each reference panorama~(see Fig.~\ref{fig:vs}~(a)). 

The points from all reference panoramas are aggregated to make one panorama using a locally weighted mixture (see Density Map in Fig.~\ref{fig:vs}~(b)). We calculate the point density for each spatial position (average number of points per pixel) of each panorama, denoted as $d_1, \dots, d_k$. For each position, the weight for view $i$ is ${\exp(\lambda_d d_i)}/{\sum_m\exp(\lambda_d d_m)}$, where $\lambda_d$ is a hyperparameter. Hence, the points in the aggregated panorama are adaptively selected from all views, rather than superimposed blindly which would expose lighting inconsistency and misalignment artifacts. 

Finally, we do a bilinear interpolation on the aggregated points in one equirectangular to reduce the empty space between rendered pixels (see Fig.~\ref{fig:vs}~(c)). 


See the first row of Fig.~\ref{fig:synth_qual} which shows the so-far output still includes major artifacts, including stitching marks, deformed objects, or large dis-occluded regions.



{\bf Neural Network Based Rendering}. 
We use a neural network, $f$ or ``filler", to fix artifacts and generate a more real looking image given the output of geometric point cloud rendering. We use a set of novelties to produce good results efficiently, including a stochastic identity initialization and adding color moment matching in perceptual loss.



\begin{figure}
	\includegraphics[width=1\linewidth]{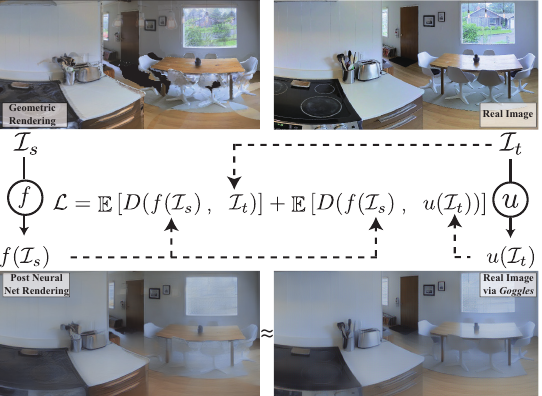}
	\caption{ \textbf{Loss configuration for neural network based view synthesis.} The loss contains two terms. The first is to transform the renderings to ground truth target images. The second is to alter ground truth target images to match the transformed rendering. A sample case is shown. (Best viewed on screen and zoomed-in.) \vspace{-4mm}}
	\label{fig:loss}
\end{figure}

{\it Architecture: } The architecture and hyperparameters of our convolutional neural network $f$ are detailed in the \suppref{supplementary material}. We utilize dilated convolutions~\cite{yu2015multi} to aggregate contextual information. We use a 18-layer network, with $3\times3$ kernels for dilated convolution layers. The maximal dilation is $32$. This allows us to achieve a large receptive field but not shrink the size of the feature map by too much. The minimal feature map size is $\frac{1}{4} \times \frac{1}{4}$ of the original image size. We also use two architectures with the number of kernels being $48$ or $256$, depending on whether speed or quality is prioritized.


{\it Identity Initialization: } Though the output of the point cloud rendering suffers from notable artifacts, it is yet quite close to the ground truth target image numerically. Thus, an identity function (i.e. input image=ouput image) is a good place for initializing the neural network $f$ at. We develop a stochastic approach to initializing the network at identity, to keep the weights nearly randomly distributed. We initialize \emph{half} of the weights  randomly with Gaussian and \emph{freeze} them, then optimize the rest with back propagation to make the network's output the same as input. After convergence, the weights are our stochastic identity initialization. Other forms of identity initialization involve manually specifying the kernel weights, e.g. \cite{chen2017fast}, which severely skews the distribution of weights (mostly 0s and some 1s). We found that to lead to slower converge and poorer results.

{\it Loss:} We use a perceptual loss~\cite{johnson2016perceptual} defined as: 
{\small
	\begin{align}
	\small
	D(I_1, I_2) &= \sum_l \lambda_l ||\Psi_l(I_1) - \Psi_l(I_2) ||_1 + \gamma \sum_{i,j} ||\bar{I}_{1_{i,j}} - \bar{I}_{2_{i,j}} ||_1. \nonumber
	\end{align} 
}%
For $\Psi$, we use a pretrained VGG16~\cite{simonyan2014very}. $\Psi_l(I)$ denotes the feature map for input image $I$ at $l$-th convolutional layer. We used all layers except for output layers. $\lambda_l$ is a scaling coefficient normalized with the number of elements in the feature map. 
We found perceptual loss to be inherently lossy w.r.t. color information (different colors were projected on one point). Therefore, we add a term to enforce matching statistical moments of color distribution. ${\bar{I}_{i,j}}$ is the average color vector of a $32\times 32$ tile of the image which is enforced to be matching between $I_1$ and $I_2$ using L1 distance and $\gamma$ is a mixture hyperparameter. We found our final setup to produce superior rendering results to GAN based losses (consistent with some recent works~\cite{chen2017photographic}).



\subsubsection{Closing the Gap with Real-World: \textbf{\emph{Goggles}} }
\label{sec:dt}
With all of the imperfections in 3D inputs and geometric renderings, it is implausible to gain fully photo-realistic rendering with neural network fixes. Thus a domain gap with real images would remain. Therefore, we instead formulate the rendering problem as forming a joint space~\cite{sener2016learning} (elaborated below) ensuring a correspondence between rendered and real images, and consequently, dissolving the gap. 

If one wishes to create a mapping $S \mapsto T$ between domain $S$ and domain $T$ by training a function $f$, usually a loss with the following form is optimized:
\begin{align}
\mathcal{L} = \mathop{\mathbb{E}}\left[D(f(\mathcal{I}_s) , \mathcal{I}_t\right)],
\end{align}
where $\mathcal{I}_s \in S, \mathcal{I}_t \in T$, and $D$ is a distance function. However, in our case the mapping between $S$ (renderings) and $T$ (real images) is not bijective, or at least the two directions $S \mapsto T$ and $T \mapsto S$ do not appear to be equally difficult. As an example, there is no unique solution to dis-occlusion filling, so the domain gap cannot reach zero exercising only $S \mapsto T$ direction. Hence, we add another function $u$ to jointly utilize $T \mapsto S$ and define the objective to be minimizing the distance between $f(\mathcal{I}_s)$ and $u(\mathcal{I}_t)$. Network $u$ is trained to alter an image taken in real-world, $\mathcal{I}_t$, to look like the corresponding rendered image in Gibson, $\mathcal{I}_s$, after passing through network~$f$ (see Fig.~\ref{fig:loss}). Function $u$ can be seen as corrective glasses of the agent, thus the name \emph{Goggles}. 

To avoid the trivial solution of all images collapsing to a single point, we add the first term in the following final loss to enforce preserving a one-to-one mapping. The loss for training networks $u$ and $f$ is:
\begin{align}
\mathcal{L} = \mathop{\mathbb{E}}\left[D(f(\mathcal{I}_s) , \mathcal{I}_t\right)] + \mathop{\mathbb{E}}\left[D(f(\mathcal{I}_s) , u(\mathcal{I}_t)\right)].
\end{align}
See Fig.~\ref{fig:loss} for a visual example. $D$ is the distance defined in Sec~\ref{subsec:viewsyn}. We use the same network architecture for $f$ and $u$. 

\subsection{Embodiment and Physics Integration}
\label{subsec:physics}

Perception and physical constraints are closely related. For instance, the perception model of a human-sized agent should seamlessly develop the notion that it does not fit in the gap under the door and hence should not attend such areas when solving a navigation task; a mouse-sized agent though could fit and its perception should attend such areas. It is thus important for the agent to be constantly subject to constraints of space and physics, e.g. collision, gravity, friction, throughout learning.

We integrated Gibson with a physics engine PyBullet \cite{coumans2018} which supports rigid body and soft body simulation with discrete and continuous collision detection.
We also use PyBullet's built-in fast collision handling system to record agent's certain interactions, such as how many times it collides with physical obstacles. 
We use Coulomb friction model by default, as scanned models do not come with material property annotations and certain physics aspects, such as friction, cannot be directly simulated. 

\begin{figure}
	\includegraphics[width=1\linewidth]{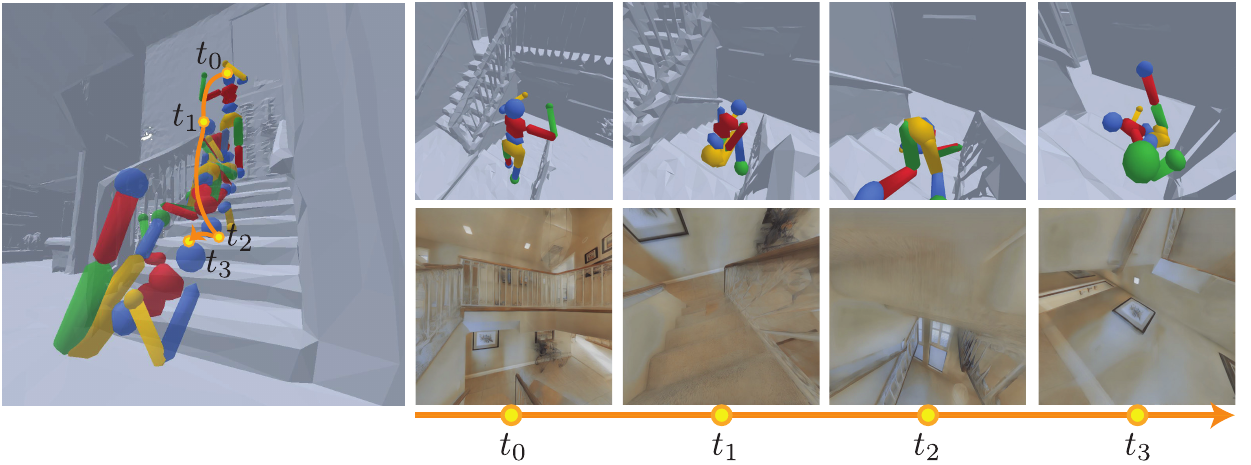}
    \vspace{-0.68cm}		
	\caption{\textbf{Physics Integration and Embodiment.} A Mujoco humanoid model is dropped onto a stairway demonstrating a physically plausible fall along with the corresponding visual observations by the humanoid's eye. The first and second rows show the physics engine view of 4 sampled time steps and their corresponding rendered RGB views, respectively.}
	\label{fig:physics}
\end{figure}

\begin{figure*}
    \vspace{-0.4cm}	
	\includegraphics[width=\linewidth]{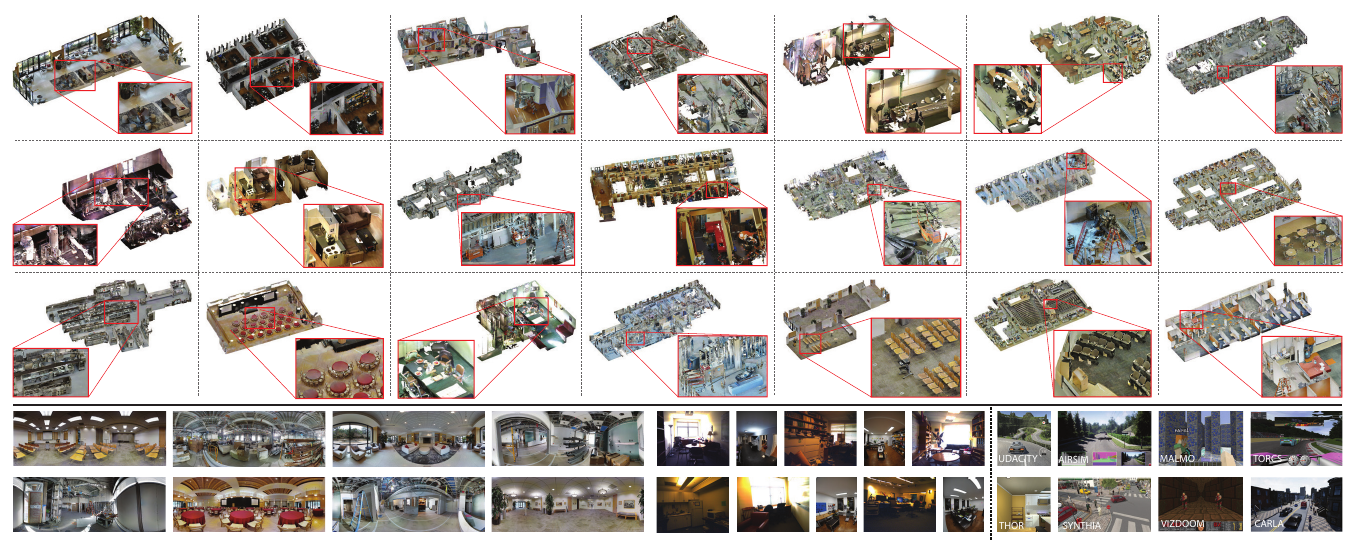}
	 \vspace{-0.8cm}	
	\caption{\textbf{Sample spaces in Gibson database.} The spaces are diverse in terms of size, visuals, and function, e.g. businesses, construction sites, houses. Upper: Sample 3D models. Lower: Sample images from Gibson database (left) and some of other environments~\cite{dosovitskiy2017carla,johnson2016malmo,RosCVPR16, airsim2017fsr,kempka2016vizdoom,wymann2000torcs,geiger2013vision, zhu2017target} (right).    \vspace{-0.2cm}		 }
	\label{fig:spaces}
\end{figure*}

\textbf{Agents:} Gibson supports importing arbitrary agents with URDFs. Also, a number of agents are integrated as entry points, including humanoid and ant of Roboschool~\cite{roboschool,2017arXiv170706347S}, husky car~\cite{Husky}, drone, minitaur~\cite{minitaur}, Jackrabbot~\cite{JR}. Agent models are in ROS or Mujoco XML format. 

\textbf{Integrated Controllers:}
To enable (optionally) abstracting away low-level control and robot dynamics for the tasks that are wished to be approached in a more high-level manner, we also provide a set of practical and ideal controllers to deduce the complexity of learning to control from scratch. We integrated a PID controller and a Nonholonomic controller as well as an ideal positional controller which completely abstracts away agent's motion dynamics.

\subsection{Additional Modalities}
Besides rendering RGB images, Gibson provides additional channels, such as depth, surface normals, and semantics. Unlike RGB images, these channels are more robust to noise in input data and lighting changes, and we render them directly from mesh files. Geometric modalities, e.g. depth, are provided for all models and semantics are available for 52,561 $m^2$ of area with semantic annotations from 2D-3D-S~\cite{2017arXiv170201105A} and Matterport3D~\cite{chang2017matterport3d} datasets.

Similar to other robotic simulation platforms, we also provide configurable proprioceptive sensory data. A typical proprioceptive sensor suite includes information of joint positions, angle velocity, robot orientation with respect to navigation target, position and velocity. We refer to this typical setup as ``non-visual sensory" to distinguish from ``visual" modalities in the rest of the paper.



\section{Tasks}
\label{sec:learn}

\textbf{Input-Output Abstraction:}
Gibson allows defining arbitrary tasks for an agent. To provide a common abstraction for this, 
we follow the interface of OpenAI Gym~\cite{brockman2016openai}: at each timestep, the agent performs an action at the environment; then the environment runs a forward step (integrated with the physics engine) and returns the accordingly rendered visual observation, reward, and termination signal. 
We also provide utility functions to keyboard operate an agent or visualize a recorded run. 

\subsection{Experimental Validation Tasks}
\label{sec:valtasks}
In our experiments, we use a set of sample active perceptual tasks and static-recognition tasks to validate Gibson. The active tasks include: 


\textbf{Local Planning and Obstacle Avoidance:} An agent is randomly placed in an environment and needs to travel to a random nearby target location provided as relative coordinates (similar to flag run~\cite{roboschool}). The agent receives no information about the environment except a continuous stream of depth and/or RGB frames and needs to plan perceptually (e.g. go around a couch to reach the target behind). 

\textbf{Distant Visual Navigation:} Similar to the the previous task, but the target location is significantly further away and fixed. Agent's initial location is still randomized. This is similar to the task of auto-docking for robots from a distant location. Agent receives no external odometry or GPS information, and needs to form a contextual map to succeed.

\textbf{Stair Climb}: An (ant~\cite{roboschool}) agent is placed on on top of a stairway and the target location is at the bottom. It needs to learn a controller for its complex dynamics to plausibly go down the stairway without flipping, using visual inputs.  

To benchmark how close to real images the renderings of Gibson are, we used two static-recognition tasks: depth estimation and scene classification. We train a neural network using  $(rendering,ground$ $truth)$ pairs as training data, but test them on $(real$ $image, ground$ $truth)$. If Gibson renderings are close enough to real images and \emph{Goggles} mechanism is effective, test results on real images are expected to be satisfactory. This also enables quantifying the impact of \emph{Goggles}, i.e. using $u(\mathcal{I}_t)$ vs. $\mathcal{I}_s, f(\mathcal{I}_s)$, and $\mathcal{I}_t$.

\textbf{Depth Estimation:} Predicting depth given a single RGB image, similar to~\cite{eigen2014depth}.
We train 4 networks to predict the depth given one of the following 4 as input images: $\mathcal{I}_s$ (pre-neural network rendering),$f(\mathcal{I}_s)$ (post-neural network rendering), $u(\mathcal{I}_t)$ (real image seen with \emph{Goggles}), and $\mathcal{I}_t$ (real image). We compare the performance of these in Sec.~\ref{sec:transfer_exp}.

\textbf{Scene Classification:} The same as previous task, but the output is scene classes rather than depth. As our images do not have scene class annotations, we generate them using a well performing network trained on Places dataset~\cite{zhou2017places}. 


\begin{figure*}
	\vspace{-0.68cm}		
	\includegraphics[width=1\linewidth]{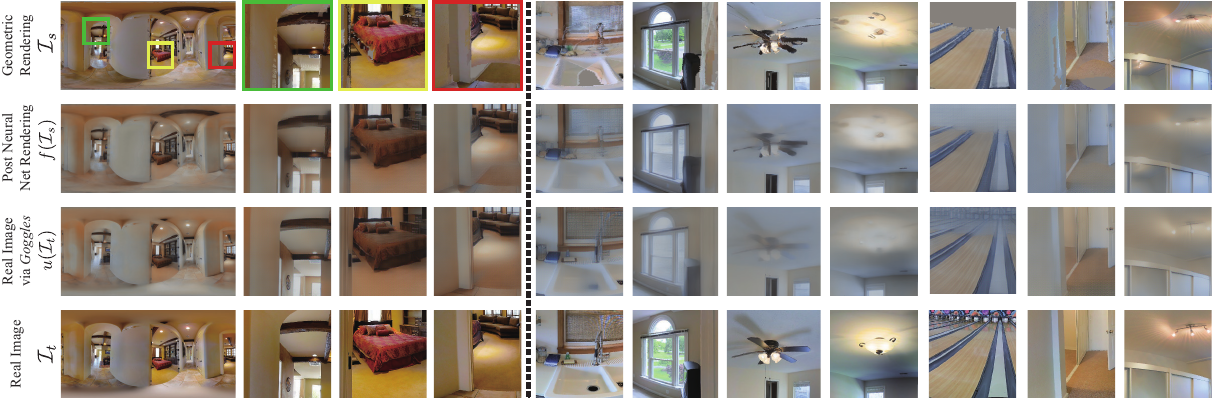}
	\vspace{-6mm}	
	\caption{\textbf{Qualitative results of view synthesis and \emph{Goggles}.} Top to bottom rows show images before neural network correction, after neural network correction, target image seen through \emph{Goggles}, and target image (i.e. ground truth real image). The first column shows a pano and the rest are sample zoomed-in patches. Note the high similarity between $2^{nd}$ and $3^{rd}$ row, signifying the effectiveness of \emph{Goggles}. (Best viewed on screen and zoomed-in.) \vspace{-4mm}}
	\label{fig:synth_qual}
\end{figure*}

\section{Experimental Results}
\label{sec:exp}

\subsection{Benchmarking Space Databases}
The spaces in Gibson database are collected using various scanning devices, including NavVis, Matterport, or DotProduct, covering a diverse set of spaces, e.g. offices, garages, stadiums, grocery stores, gyms, hospitals, houses. All spaces are fully reconstructed in 3D and post processed to fill the holes and enhance the mesh. We benchmark some of the existing synthetic and real databases of spaces (SUNCG~\cite{song2016ssc} and Matterport3D~\cite{chang2017matterport3d}) vs Gibson's using the following metrics in Table~\ref{tbl:modelcomp}: 

{\bf Specific Surface Area (SSA)}: the ratio of inner mesh surface and volume of convex hull of the mesh. This is a measure of clutter in the models.

{\bf Navigation Complexity}: Longest $A^*$ navigation distance between randomly placed two points divided by the straight line distance. We compute the highest navigation complexity $\max_{s_i, s_j} \frac{d_{A^*}(s_i, s_j)}{d_{l2}(s_i, s_j)}$ for every model.

{\bf Real-World Transfer Error:} We train a neural network for depth estimation using the images of each database and test them on real images of 2D-3D-S dataset~\cite{2017arXiv170201105A}. Training images of SUNCG and Matterport3D are rendered using MINOS~\cite{savva2017minos} and our dataset is rendered using Gibson's engine. The training set of each database is 20k random RGB-depth image pairs with $90^\circ$ field of view. The reported value is average depth estimation error in meters. 

{\bf Scene Diversity}: We perform scene classification on 10k randomly picked images for each database using a network pretrained on~\cite{zhou2017places}. We report the entropy of the distribution of top-1 classes for each environment. Gibson, SUNCG~\cite{song2016ssc}, and THOR~\cite{zhu2017target} gain the scores of $3.72$, $2.89$, and $3.32$, respectively (highest possible entropy = $5.90$). \vspace{-7mm}

\begin{flushleft}
	\begin{table}
		\small
		\centering
		\begin{tabular}{l{c}{c}{c}}
			Dataset           & Gibson & SUNCG & Matterport3D  \\
			\hline
			{Number of Spaces} & 572 & 45622  & 90\\
			{Total Coverage~$m^2$}  & 211k & 5.8M & 46.6K\\
			{SSA} & 1.38 & 0.74 & 0.92\\ 
			{Nav. Complexity} & 5.98 & 2.29 & 7.80\\
			{Real-World Transfer Err} & 0.92$^\S$ & 2.89$^\dagger$ & 2.11$^\dagger$\\
		\end{tabular}
		\vspace{-0.2cm}		
		\caption{\textbf{Benchmarking Space Databases:} Comparison of Gibson database with SUNCG~\cite{song2016ssc} (hand designed synthetic), and Matterport3D~\cite{chang2017matterport3d}. $^\S$ Rendered with Gibson, $^\dagger$ rendered with MINOS~\cite{savva2017minos}.\vspace{-3mm}}
		\label{tbl:modelcomp}
	\end{table}
\end{flushleft}


\subsection{Evaluation of View Synthesis}
To train the networks $f$  and $u$ of our neural network based synthesis framework, we sampled 4.3k $1024\times 2048$ $\mathcal{I}_s$|$\mathcal{I}_t$ panorama pairs and randomly cropped them to $256\times 256$. We use Adam~\cite{kingma2014adam} optimizer with learning rate $2\times 10 ^{-4}$. We first train $f$ for 50 epochs until convergence, then we train $f$ and $u$ jointly for another 50 epochs with learning rate $2\times 10^{-5}$. The learning finishes in 3 days on 2 Nvidia Titan X GPUs. 

Sample renderings and their corresponding real image (ground truth) are shown in Fig.~\ref{fig:synth_qual}. Note that pre-neural network renderings suffer from geometric artifacts which are partially resolved in post-neural network results. Also, though the contrast of the post-neural network images is lower than real ones and color distributions are still different, \emph{Goggles} could effectively alter the real images to match the renderings (compare $2^{nd}$ and $3^{rd}$ rows). In additional, the network $f$ and \emph{Goggles} $u$ jointly addressed some of the pathological domain gaps. For instance, as lighting fixtures are often thin and shiny, they are not well reconstructed in our meshes and usually fail to render properly. Network $f$ and \emph{Goggles} learned to just suppress them altogether from images to not let a domain gap remain. The scene out the windows also often have large re-projection errors, so they are usually turned white by $f$ and $u$. 

 Appearance columns in Table~\ref{tbl:transfer} quantify view synthesis results in terms image similarity metrics L1 and SSIM. They echo that the smallest gap is between $f(\mathcal{I}_s)$ and $u(\mathcal{I}_t)$.

\textbf{Rendering Speed} of Gibson is provided in Table~\ref{fig:efficiency}.

\begin{table}
	\small
	\begin{tabular}{l|*{3}{c}r}
		Resolution              & 128\small{x}128 & 256\small{x}256 & 512\small{x}512  \\
		\hline
		RGBD, pre  network$f$ & 109.1 & 58.5 & 26.5 \\
		RGBD, post network$f$ & 77.7 & 30.6 & 14.5 \\
		RGBD, post small network$f$ & 87.4 & 40.5 & 21.2 \\
		Depth only & 253.0 & 197.9 & 124.7 \\
		Surface Normal only & 207.7 & 129.7 & 57.2 \\
		Semantic only & 190.0 & 144.2 & 55.6 \\
		Non-Visual Sensory  & 396.1 & 396.1 & 396.1
	\end{tabular}
	\vspace{-0.2cm}		
	\caption{\textbf{Rendering speed (FPS)} of Gibson on a single GPU for different resolutions and output configurations. Tested on E5-2697 v4 with Tesla V100 in headless rendering mode. As a faster setup (``small network"), we also trained a smaller filler network with donwsized input geometric renderings. This setup achieves a higher FPS at the expense of inferior visual quality compared to full-size filler network.\vspace{-3mm}}
	\label{fig:efficiency}
\end{table}

\begin{table}
	
	\centering
	\begin{tabular}{l|l|p{1.2cm} p{1.2cm} |ll}
		\multirow{2}{*}{Train} & \multirow{2}{*}{Test} &  \multicolumn{2}{c|}{Static Tasks} &  \multicolumn{2}{c}{Appearance} \\
		\hhline{~~----}
		& & \footnotesize{Scene Class Acc.}  & \footnotesize{Depth Est. Error}  & SSIM & L1   \\
\hline
$\mathcal{I}_s$ & $\mathcal{I}_t$ & 0.280 & 1.026  & 0.627 & 0.096   \\
$f(\mathcal{I}_s)$ & $\mathcal{I}_t$  & 0.266 & 1.560  & 0.480 & 0.10   \\
$f(\mathcal{I}_s)$ & $u(\mathcal{I}_t)$ & \bf{0.291} & \bf{0.915}  & \bf{0.816} & \bf{0.051}    
	\end{tabular}
	\caption{\textbf{Evaluation of view synthesis and transferring to real-world.} \emph{Static Tasks} column shows on both scene classification task and depth estimation tasks, it is easiest to transfer from $f(\mathcal{I}_s)$ to $u(\mathcal{I}_t)$ compared with other cross-domain transfers. \emph{Appearance} columns compare L1 and SSIM distance metrics for different pairs showing the combination of network $f$ and \emph{Goggles} $u$ achieves best results.\vspace{-5mm}}
	\label{tbl:transfer}
\end{table}

\subsection{Transferring to Real-World}
\label{sec:transfer_exp}

We quantify the effectiveness of \emph{Goggles} mechanism in reducing the domain gap between Gibson renderings and real imagery in two ways: via the static-recognition tasks described in Sec.~\ref{sec:valtasks} and by comparing image distributions. 

\begin{figure}
    \vspace{-0.2cm}		
	\centering
	{\includegraphics[width=1\linewidth]{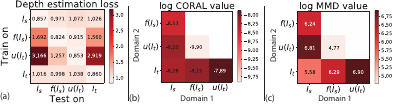}}
	\caption{\textbf{Evaluation of transferring to real-world from Gibson.} (a) Error of depth estimation for all train-test combinations. (b,c) MMD and CORAL distributional distances. All tests are in support of \emph{Goggles}.}
	\label{fig:domain_adapt_res}
\end{figure}

Evaluation of transferring to real images via scene classification and depth estimation are summarized in Table.~\ref{tbl:transfer}. Also, Fig.~\ref{fig:domain_adapt_res}~(a) provides depth estimation results for all feasible train-test combinations for reference. The diagonal values of the $4\times 4$ matrix represent training and testing on the same domain. The gold standard is train and test on $\mathcal{I}_t$ (real images) which yields the error of 0.86. The closest combination to that in the entire table is train on $f(I_s)$ ($f$ output) and test on $u(I_t)$ (real image through \emph{Goggles}) giving 0.91, which signifies the effectiveness of \emph{Goggles}.  


In terms of distributional quantification, we used two metrics of Maximum Mean Discrepancy (MMD)~\cite{gretton2012kernel} and CORAL~\cite{sun2016deep} to test how well $f(\mathcal{I}_s)$ and $u(\mathcal{I}_t)$ domains are aligned. The metrics essentially determine how likely it is for two samples to be drawn from different distributions. We calculate MMD and CORAL values using the features of the last convolutional layer of VGG16~\cite{simonyan2014very} and kernel $k(x,y) = x^Ty$. Results are summarized in Fig.~\ref{fig:domain_adapt_res}~(b) and (c). For each metric, $f(\mathcal{I}_s)$ - $u(\mathcal{I}_t)$ is smaller than other pairs, showing that the two domains are well matching. 

In order to quantitatively show the networks $f$ and $u$ do not give degenerate solutions (i.e. collapsing all images to few points to close the gap by cheating), we use $f(\mathcal{I}_s)$ and $u(\mathcal{I}_t)$ as queries to retrieve their nearest neighbor using VGG16 features from $\mathcal{I}_s$ and $\mathcal{I}_t$, respectively. Top-1, 2 and 5 accuracies for $f(\mathcal{I}_s)\mapsto\mathcal{I}_s$ are 91.6\%, 93.5\%, 95.6\%. Top-1, 2 and 5 accuracies for $u(\mathcal{I}_t)\mapsto\mathcal{I}_t$ are 85.9\%, 87.2\%,89.6\%. This indicates a good correspondence between pre and post neural network images is preserved, and thus, no collapse is observed.

\subsection{Validation Tasks Learned in Gibson}
\vspace{-1mm} 
The results of the active perceptual tasks discussed in Sec.~\ref{sec:valtasks} are provided here. 
In each experiment, the non-visual sensor outputs include agent position, orientation, and relative position to target. The agents are rewarded by the decrease in their distance towards their targets. In Local Planning and Visual Obstacle Avoidance, they receive an additional penalty for every collision.

\begin{figure}
	\includegraphics[width=0.45\linewidth]{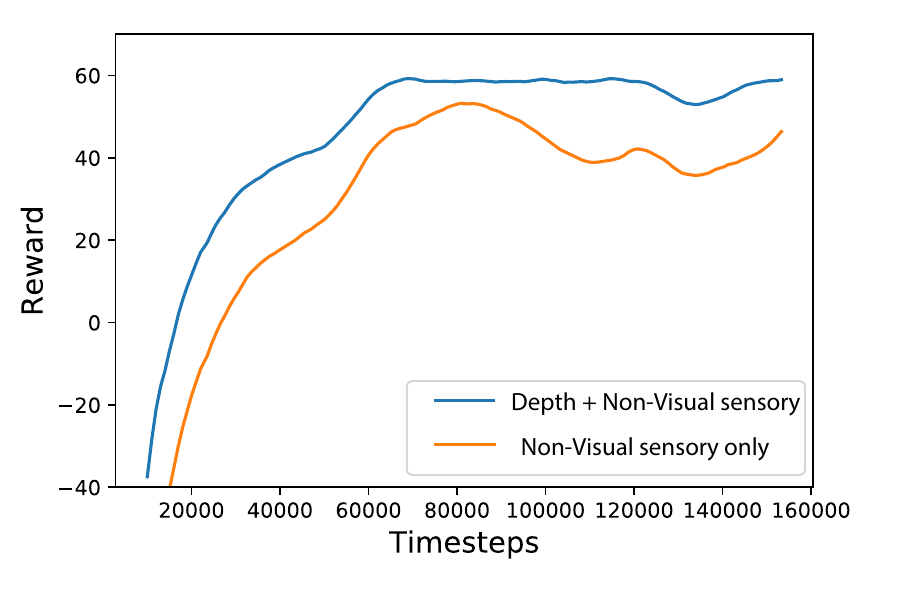}
	\includegraphics[width=0.45\linewidth]{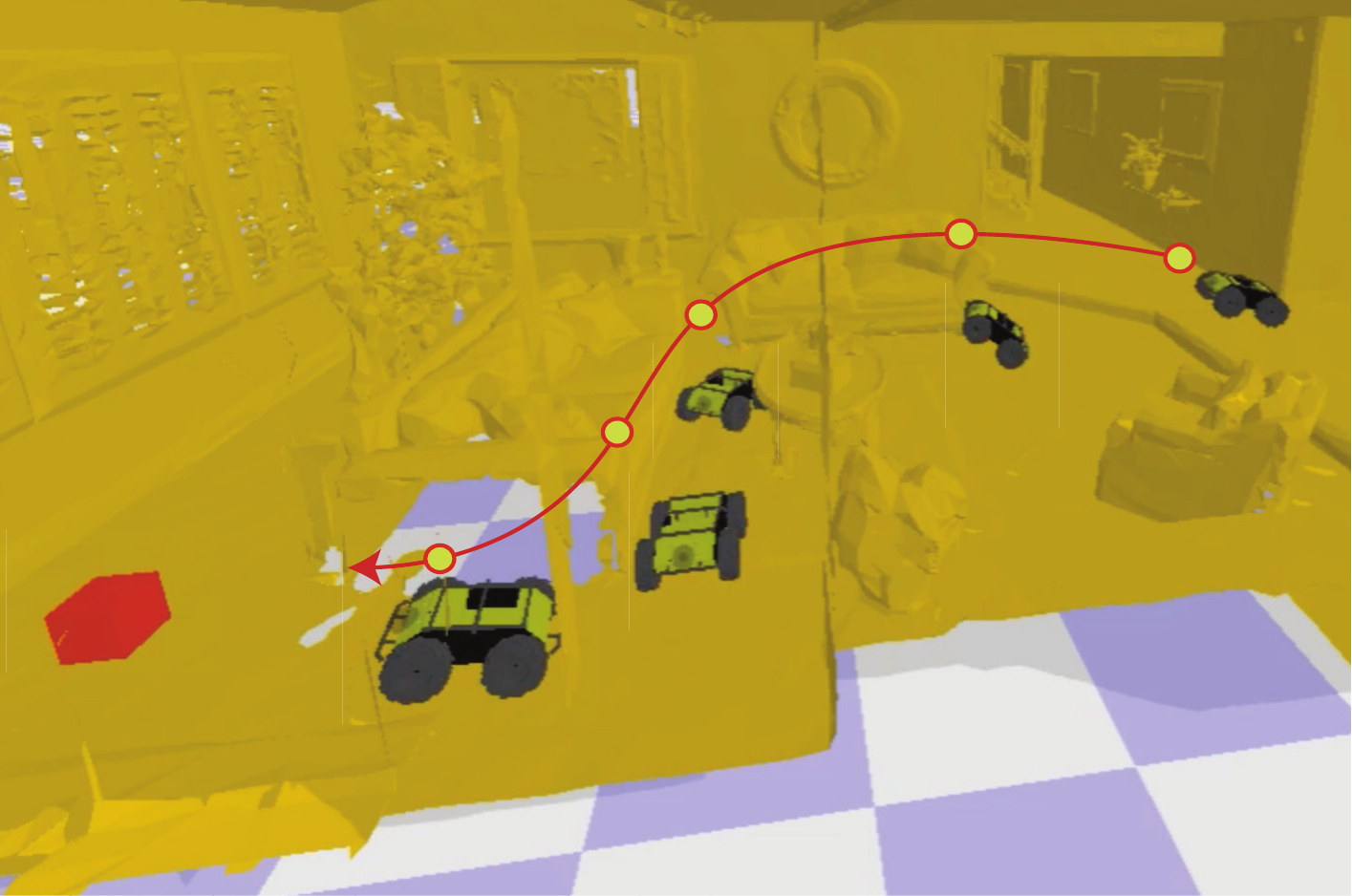}
	\caption{\textbf{Visual Local planning and obstacle avoidance.} Reward curves for perceptual vs non-perceptual husky agents and a sample trajectory.\vspace{-0.2cm}}
	\label{fig:reward_flagrun}
\end{figure}

{\bf Local Planning and Visual Obstacle Avoidance Results:}
We trained a perceptual and non-perceptual husky agent according to the setting in Sec.~\ref{sec:valtasks} with PPO~\cite{schulman2017proximal} for 150 episodes (300 iterations, 150k frames). Both agents have a four-dimensional discrete action space: forward/backward/left/right. The average reward over 10 iterations are plotted in Fig~\ref{fig:reward_flagrun}. The agent with perception achieves a higher score and developed obstacle avoidance behavior to reach the goal faster.

\begin{figure}
    \vspace{-0.2cm}			
	\includegraphics[width=1\linewidth]{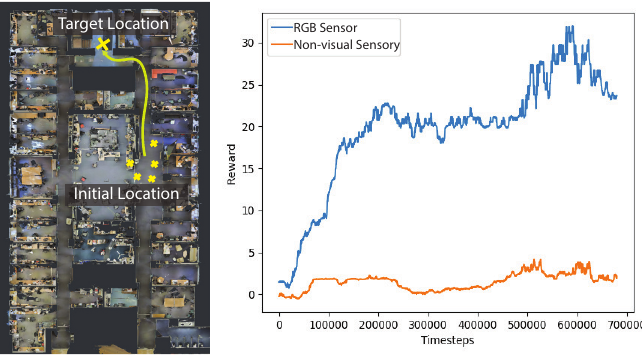}
    \vspace{-0.5cm}				
	\caption{\textbf{Distant Visual Navigation.} The initial locations and target are shown. The agent succeeds only when provided with visual inputs.    \vspace{-0.5cm} }
	\label{fig:reward_navigation}
\end{figure}

{\bf Distant Visual Navigation Results:} 
Fig.~\ref{fig:reward_navigation} shows the target and sample random initial locations as well as the reward curves. Global navigation behavior emerges after 1700 episodes (680k frames), and only the agent with visual state was able to accomplish the task. The action space is the same as previous experiment.

Also, we use the trained policy of distant navigation to evaluate the impact of \emph{Goggles} on an active task: we go to camera locations where $\mathcal{I}_t$ is available. Then we measure the policy discrepancy in terms of L2 distance of output action logits when different renderings and $\mathcal{I}_t$ are provided as input. Training on $f(\mathcal{I}_s)$ and testing on $u(\mathcal{I}_t)$ yields discrepancy of 0.204 (best), while training on $f(\mathcal{I}_s)$ and testing on $\mathcal{I}_t$ gives 0.300 and training on  $\mathcal{I}_s$ and testing on $\mathcal{I}_t$ gives 0.242. After the initial release of our work, a paper recently reported an evaluation done on a real robot for adaptation using backward mapping from real images to renderings~\cite{zhang2018vr}, with positive results. They did not use paired data, unlike Gibson, which would be expected to further enhance the results.

{\bf Stair Climb:} As explained in Sec.~\ref{sec:valtasks}, an ant~\cite{roboschool} is trained to perform the complex locomotive task of plausibly climbing down a stairway without flipping. The action space is eight dimensional continuous torque values. We train one perceptual and one non-perceptual agent starting at a fixed initial location, but at test time slightly and randomly move their initial and target location around. They start to acquire stair-climbing skills after 1700 episodes (700k time steps). While the perceptual agent learned slower, it showed better generalizability at test time coping with the location shifts and outperformed the non-perceptual agent by 70\%. Full details of this experiment is privded in the \suppref{supplementary material}.


\vspace{-1mm} 
\section{Limitations and Conclusion}
\label{sec:conclution}
\vspace{-2mm}

We presented Gibson Environment for developing real-world perception for active agents and validated it using a set of tasks. While we think this is a step forward, there are some limitations that should be noted. First, though Gibson provides a good basis for learning complex navigation and locomotion, currently it does not include dynamic content (e.g. other moving objects) and does not support manipulation. This can be potentially solved by integrating our approach with synthetic objects~\cite{chang2015shapenet,karsch2011rendering}. Second, we do not have full material properties and no existing physics simulator is optimal; this may lead to physics related domain gaps. Finally, we provided quantitative evaluations of \emph{Goggles} mechanism for transferring to real world mostly using static recognition tasks. The ultimate test is evaluating \emph{Goggles} on real robots.
\\ \\
\noindent\textbf{Acknowledgement:} We gratefully acknowledge the support of Facebook, Toyota (1186781-31-UDARO), ONR MURI (N00014-14-1-0671), ONR (1165419-10-TDAUZ); Nvidia, CloudMinds, Panasonic (1192707-1-GWMSX).

{\small
\bibliographystyle{ieee}
\bibliography{reference}
}

\clearpage
\makeatletter
\let\maketitle\suppsavedmaketitle
\let\@maketitle\suppsavedatmaketitle
\let\title\suppsavedtitle
\let\author\suppsavedauthor
\let\and\suppsavedand
\let\thanks\suppsavedthanks
\let\date\suppsaveddate
\makeatother
\renewcommand{\baselinestretch}{1}\selectfont
\setcounter{page}{1}
\setcounter{footnote}{0}
\setcounter{section}{0}
\setcounter{figure}{0}
\setcounter{table}{0}
\setcounter{equation}{0}
\renewcommand{\theHsection}{supp.\arabic{section}}
\renewcommand{\theHfigure}{supp.\arabic{figure}}
\renewcommand{\theHtable}{supp.\arabic{table}}
\renewcommand{\theHequation}{supp.\arabic{equation}}
\def\theHfootnote{supp.\arabic{footnote}}

\title{Gibson Env: Real-World Perception for Embodied Agents \\ \large{\underline{Supplementary Material}}}

\author{Fei Xia\thanks{Authors contributed equally.}$_{1}$ \hspace{1mm} \;\;Amir R. Zamir\footnotemark[1]$_{1,2}$ \hspace{1mm} \;\;Zhiyang He\footnotemark[1]$_{1}$  \;\; Alexander Sax$_{1}$  \;\; Jitendra Malik$_{2}$ \;\; Silvio Savarese$_{1}$\vspace{5pt}\\ 
	$^1$ Stanford University  \;\;  
	$^2$ University of California, Berkeley\vspace{5pt}\\ 
	{\url{http://gibsonenv.stanford.edu/}\vspace{0pt}}
}

\maketitle
\hypertarget{supplement.start}{}

\begin{abstract}
The following items are provided in the supplementary material:
\begin{enumerate}[nosep]
    \item Details of point cloud interpolation and view selection in view synthesis, and details of network architecture and hyper parameters. 
    \item More information about semantic modality in our environment. 
    \item Additional view synthesis qualitative results.
    \item RL training details and additional RL results in our environment.    
\end{enumerate}

\end{abstract}


\section{View Synthesis Details}
\subsection{Details of Point Cloud Rendering}
Here we provide more details about the point cloud interpolation and view selection part of view synthesis. For a view $v_j$ that we want to synthesis, we get the rendered point cloud from the nearest $k$ views $v_{j1}, v_{j2}, \dots, v_{jk}$ based on Euclidean distance. Denote the reprojected point clouds as $p_{j1}, p_{j2}, \dots, p_{jk}$. Assuming the target resolution is $w\times h$, each point cloud is a set of projected points $\{(x,y)\}$, with $0\leq x \leq w$ and $0 \leq y \leq h$. We do two operations on each project point cloud. First, we run a kernel density estimator on the point cloud with a Gaussian kernel. This provides information about the density of points from a certain view at a certain region. Second, for each grid point in $x,y$-plane we query the nearest 4 points and did a bilinear sampling in an irregular polygon. Denote the density of point cloud image of view $k$ as $d_{jk}\in R^{w\times h}$. The weight for view $k$ is $w_{jk} = d_{jk}/ (\sum_i d_{ji})$. Denote the interpolated image from source view $k$ as $I_{jk} \in R^{w\times h\times 3}$, then the final output of view selection and interpolation is $I_j = \sum_i w_{ji} \circ I_{ji}$. We implement these steps in CUDA and thus they have small effects on point cloud rendering speed. The benefit of the point cloud interpolation and view selection is that they create a smooth transition across point clouds from different source views, making the imperfection easier to fix by Neural Network Filler $f$.

\subsection{Details of Training Neural Network Filler}
\textbf{Network Architecture.} The details of the network architecture are shown in Table~\ref{tbl:archiecture}. We use a combination of convolutional layers, deconvolutional layers and dilated convolutional layers. Batch normalization is applied to each output feature map of all types of convolutional layers. We use leaky ReLU with slope $0.1$ as activation function for all layers. 

\textbf{Identity initialization.}
For image processing or filtering, [11] advocates using adaptive normalization for identity initialization, where they also design kernels to force identity mapping. Empirically, we find that using a stochastic version that ensures an overall identity function but preserves the diversity of Gaussian random initialization works better and achieves faster convergence. We initialize half of the weight with Gaussian random and \emph{freeze} them and optimize the rest with back propagation so the network outputs the same input image. After convergence, the weights are our stochastic identity initialization.

\textbf{Hyperparameter setting.} For the network architecture, the number of kernels of the network is configurable with a parameter $n_f$. In the experiments, we use two settings where $n_f$ is either 24 or 128. For domain adaptation and view synthesis experiments we use $n_f = 128$. For RL training we use $n_f = 24$ to reduce RL training time. For the loss, there are two hyperparameters to tune. The first is $\lambda_l$, which is the coefficient for each layer of VGG feature map. 
We normalize it the number of elements in the feature map, i.e. $\lambda_l = 1/n_l$. The second parameter is the coefficient for color match loss $\gamma$. We set it to $0.05$ across the experiments.  

\begin{figure*}
	\centering
	\includegraphics[width=\linewidth]{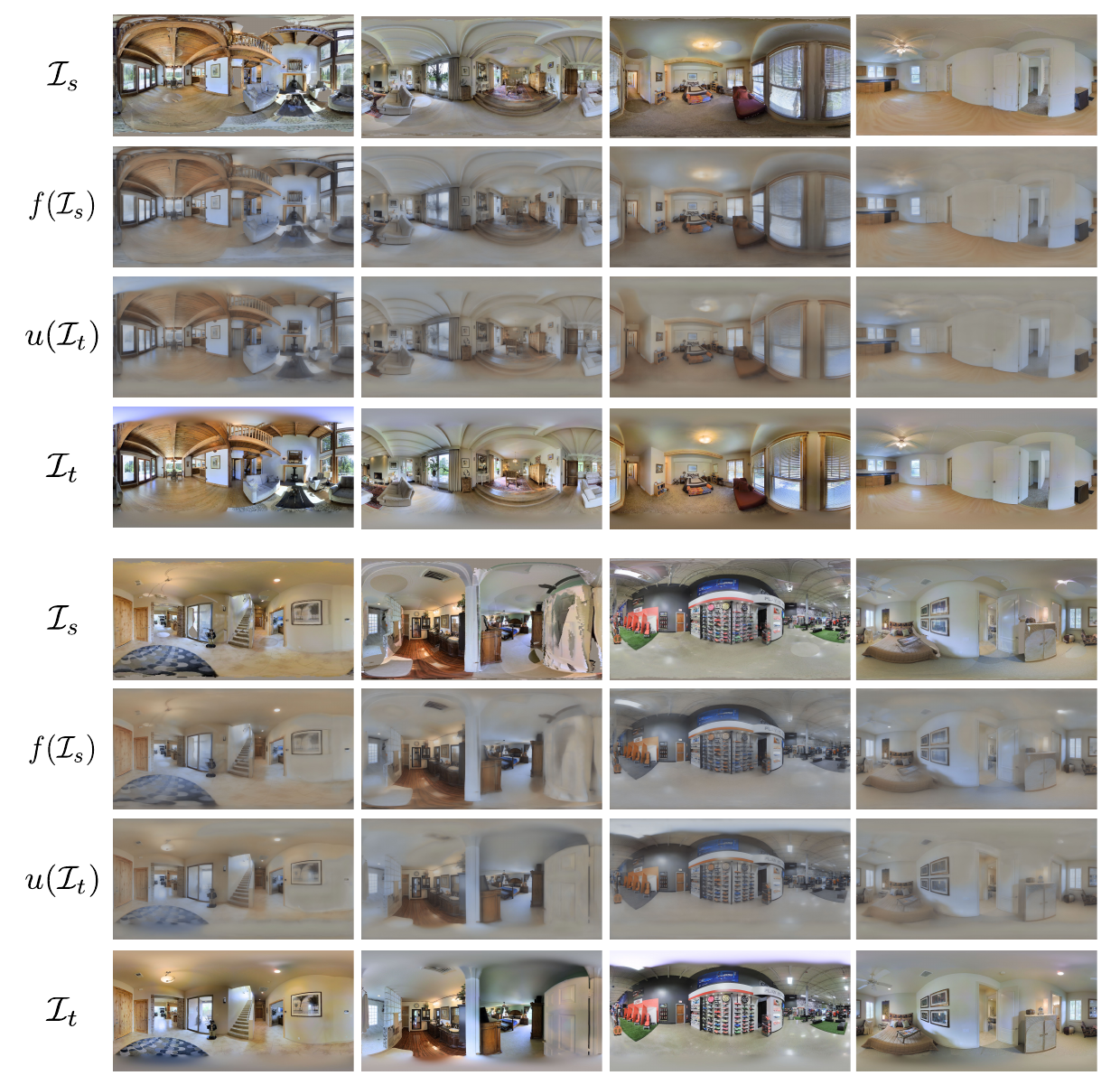}
	\caption{\textbf{Additional view synthesis results.} We provide additional view synthesis and domain adaptation results. The format is similar to Fig. 6 in the main paper. The results have high resolution and are very rich in details, so we encourage readers to zoom in to see details. }
	\label{fig:additional_vs}
\end{figure*}

\begin{table}
\centering

\begin{tabular}{llllll}
\hline 
Type   & Kernel & Dilation & Stride & Output \\
\hline 
conv.   & $5\times 5$  & 1 &   $1\times1$  &    6    \\
conv.   & $5\times 5$  & 1 &   $2\times2$  &    $n_f$    \\
conv.   & $3\times 3$  & 1 &   $1\times1$  &    $n_f$    \\
conv.   & $5\times 5$  & 1 &   $2\times2$  &    $4n_f$    \\
\hline
dilated conv.   & $3\times 3$  & 1  &   $1\times1$  &    $4n_f$    \\
dilated conv.   & $3\times 3$  & 1  &   $1\times1$  &    $4n_f$    \\
dilated conv.   & $3\times 3$  & 2  &   $1\times1$  &    $4n_f$    \\
dilated conv.   & $3\times 3$  & 4  &   $1\times1$  &    $4n_f$    \\
dilated conv.   & $3\times 3$  & 8  &   $1\times1$  &    $4n_f$    \\
dilated conv.   & $3\times 3$  & 16  &   $1\times1$  &    $4n_f$    \\
dilated conv.   & $3\times 3$  & 32  &   $1\times1$  &    $4n_f$    \\
dilated conv.   & $3\times 3$  & 1  &   $1\times1$  &    $4n_f$    \\
dilated conv.   & $3\times 3$  & 1  &   $1\times1$  &    $4n_f$    \\
\hline

deconv.   & $4\times 4$  & 1  &   $2\times2$  &    $n_f$    \\
conv.   & $3\times 3$  & 1 &   $1\times1$  &    $n_f$    \\
deconv.   & $4\times 4$  & 1  &   $2\times2$  &    $6$    \\
conv.   & $3\times 3$  & 1 &   $1\times1$  &    $6$    \\
conv.   & $3\times 3$  & 1 &   $1\times1$  &    $3$    \\

\hline

\end{tabular}
\caption{\textbf{Network Architecture.} The network architecture is detailed in this table. We use three types of layers: convolutional layers(conv.), dilated convolutional layers(dilated conv.) and deconvolutional layers(deconv.). The network size is configurable with parameter $n_f$.}
\label{tbl:archiecture}
\end{table}

\section{Details about Semantic Modality}
\begin{figure}
\includegraphics[width=1.0\linewidth]{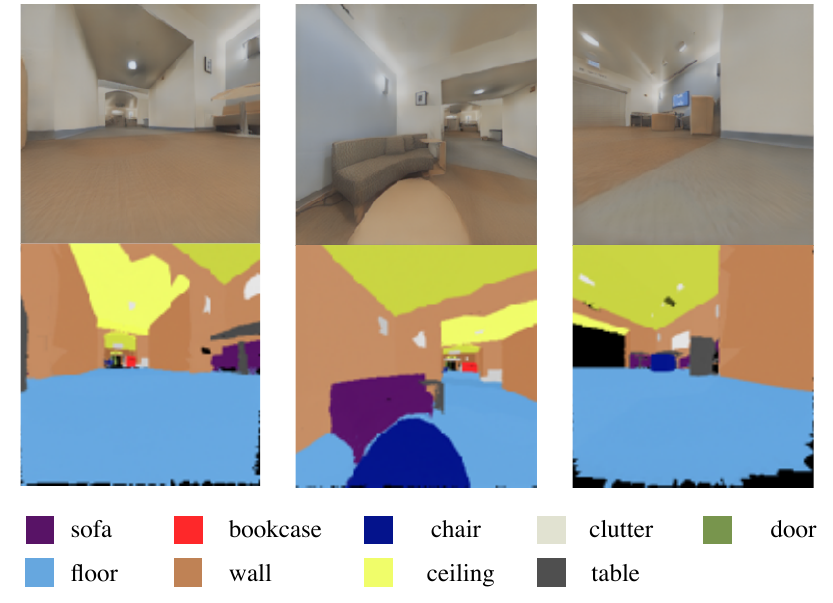}
\caption{\textbf{Sample frames with semantic annotation}. Top: NN filled images; Bottom: Corresponding semantic labels of pixels. }

\label{fig:semantics}
\end{figure}

Some models in our system are semantically annotated. For those models, we are able to output a semantic segmented frame. Examples are shown in Fig.~\ref{fig:semantics}. We provide semantic labels for 13 classes, including floor, ceiling, wall, beam, window, column, door, table, chair, bookcase, sofa, board, and clutter. 

\section{Additional Experiments}

\subsection{Additional View Synthesis Results}

We provide additional view synthesis results to provide a more thorough understanding of the qualitative results. The results are shown in Fig.~\ref{fig:additional_vs}.

\subsection{Additional RL Experiment Setup}

We used Gibson environment with default physical settings (gravity = 9.8$m^2/s^2$, friction coefficient = 0.7) for our experiments. In Visual Obstacle Avoidance and Visual Navigation we used default discrete action space in Gibson for Husky robot, which translates four dimensional (left/right/forward/backward) control signal to four sets of predefined torques. In Visuomotor Control we used default continuous action space for Ant robot, which accepts eight real value torque signals. 

At every time step, the nonviz sensor output received by the agent includes robot position, orientation, velocity, and distance and angle towards target. In Visuomotor Control experiment, this also includes the number of feed the Ant robot has in contact with the ground.

In Visual Obstacle Avoidance, Navigation and Visuomotor Control tasks, our primary reward for the agent is the change in potential (negative distance towards the target), specifically $-\frac{d L_{target}}{dt}$. In visual obstacle avoidance we added another distance term $c * L_{nearest}$, where $L_{nearest}$ is the length of the beam coming out from the front of the robot. This mimics the effect of front LiDAR and encourages the agent to move in unobstructed directions. We didn't use any additional term in our reward: alive score, collision, contact, etc. The agent learns to utilize it's DoF, minimize collisions and maximize its lifespan to reach the highest score.

\subsection{Additional RL Experiment Results}

{\bf Network Architecture and Training Details} In order to account for the different input modalities of inputs, we use different Neural Network policies. For sensor-only RL agents, we used a policy with two identical networks for value and policy function: MLP with two hidden layers using $tanh$ activation. For perceptual agents, we used a sensor-vision policy that has the same MLP structure to process sensor input, and another CNN with two hidden layers using $relu$ activation to process camera input. We use a densely connected layer to combine MLP and CNN at the end. The implementations are based on OpenAI baselines.

We used Proximal Policy Optimization for training our RL policies. We used linearly decreasing learning rate, gamma of $0.99$ and optimization batch size of $64$. Our average episode length is 400 for Navigation and 60 for Visuomotor Control. We set time step per actor batch for these two tasks as 3000 and 2000. Since there is no ending condition for Obstacle Avoidance task, we set its time step per batch as 1024. Our reward curves for Obstacle Avoidance, Navigation, and Visuomotor Control stabilize after 100K, 600K and 700K time steps.

\begin{figure}
\includegraphics[width=1.0\linewidth]{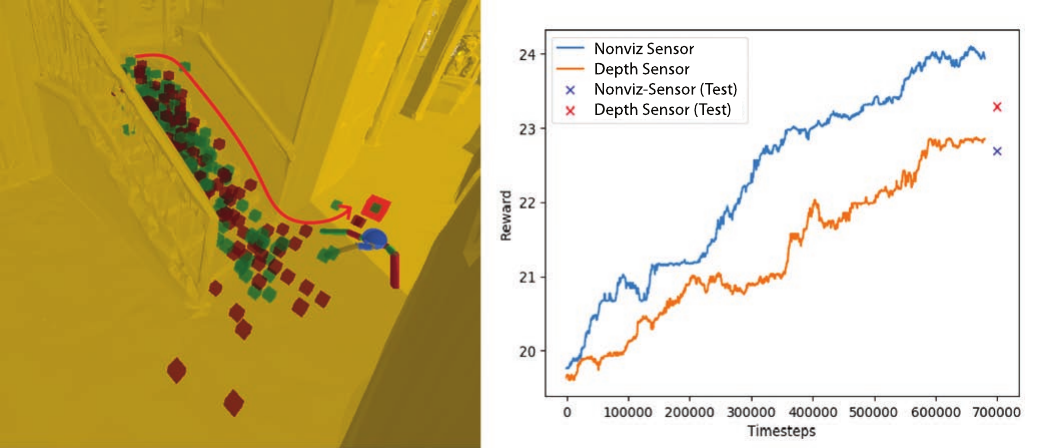}
\caption{\textbf{Ant climbing downstairs tasks reward curve}. Left: ending position of sensor-only (red) and perceptual (green) agent. Right: reward curves and test score. Here we plot the reward curve over the first 700K frames when the agent starts to acquire stair climbing ability.}

\label{fig:reward_climb}
\end{figure}

{\bf Visuomotor Control Results} During training, we modified the ant robot's size and body shape to be reasonable based on the size of stairs in real life. We initialize the ant at random location at the midpoint of the stairs and train it to reach the bottom of the stairs using the sensor-only network, and sensor-vision fusion network with depth as perceptual input. Both agents know their relative location towards the target. To prevent the agent from learning suboptimal policies (e.g. rolling down the stairs), we terminate the agent when its leaning angle is more than 90 degrees.

Within the same amount of clock time, we trained the sensor-only agent for 4M frames, and perceptual agent for 900K frames, due to longer CNN processing time in fusion network. Sensor-only agent reached a final score of 34 and perceptual agent reached a final score of 25.6.

To evaluate how robust our learned visuomotor control policies are, we modified the target positions in test time. In Fig.~\ref{fig:reward_climb} the target (light red cube) is moved from the bottom of stairs to behind the corner by 0.5$m$. In order to reach it, the agent needs to first climb the stairs and then walk towards the target. Red and green small cubes show the death/finish locations of sensor-only and perceptual agents. Performance of sensor only agent drops significantly from 34.0 to 22.7 (-11.3), while the perceptual agent's performance slightly drops from 25.6 to 23.3. Failure cases for the sensor-only robot are: (1) being obstructed by the wall, (2) not approaching the target after going downstairs. This shows that policy learned by the sensor-only robot is limited in either only knowing to move in target direction, or only knowing to climb downstairs.

\end{document}